\RequirePackage{fix-cm}
\documentclass[twocolumn]{svjour3}          
\smartqed  
\usepackage[style=authoryear-icomp,natbib=true,backend=biber]{biblatex}
\usepackage{graphicx}
\usepackage{tikz}
\usepackage{multirow}
\usepackage{pgfplots}
\pgfplotsset{compat=1.8}
\usepackage{pgfplotstable}
\usepackage[export]{adjustbox}
\usepackage{amsmath,amssymb,amsfonts}
\usepackage{algorithmic}
\usepackage{indentfirst}
\usepackage{sfmath}
\usepackage{mathdots}
\usepackage{multirow}
\usepackage{yhmath}
\usepackage{color}
\usepackage{hyperref}
\usepackage{xcolor}
\usepackage{wrapfig}

\definecolor{unlabeled}{rgb}{0.0392,    0.0392,    0.0392}
\definecolor{sky}{rgb}{0.5451,    0.6471,    0.5137}
\definecolor{water}{rgb}{0.3235,    0.5980,    0.73536}
\definecolor{windows}{rgb}{0.7255,    0.7137,    0.4471}
\definecolor{road}{rgb}{0.7216,    0.5804,    0.3412}
\definecolor{car}{rgb}{0.6412,    0.3431,    0.3039}
\definecolor{buildings}{rgb}{0.2980,    0.2980,    0.2980}
\definecolor{none}{rgb}{0.7,0.7,0.7}

\bibliography{refs.bib}
\begin{document}

\title{Towards urban scenes understanding through polarization cues
}
\subtitle{}


\author{Marc Blanchon$^1$         \and
        D{\'e}sir{\'e} Sidib{\'e}$^2$ \and
        Olivier Morel$^1$  \and
        Ralph Seulin$^1$ \and
        Fabrice Meriaudeau$^1$
}

\authorrunning{Marc Blanchon et al.} 

\institute{Marc Blanchon \at
              \email{fr.marc.blanchon@gmail.com} \and
              $^{1}$EMR VIBOT CNRS 6000, ImViA, Universit\'e Bourgogne Franche-Comt\'e, 71200, Le Creusot, France\\
$^{2}$IBISC, Univ Evry, Universit\'e Paris-Saclay, 91025, Evry, France
}

\date{Received: date / Accepted: date}

\maketitle

\begin{abstract}
Autonomous robotics is critically affected by the robustness of its scene understanding algorithms. We propose a two-axis pipeline based on polarization indices to analyze dynamic urban scenes. As robots evolve in unknown environments, they are prone to encountering specular obstacles. Usually, specular phenomena are rarely taken into account by algorithms which causes misinterpretations and erroneous estimates.
By exploiting all the light properties, systems can greatly increase their robustness to events. In addition to the conventional photometric characteristics, we propose to include polarization sensing. 

We demonstrate in this paper that the contribution of polarization measurement increases both the performances of segmentation and the quality of depth estimation. Our polarimetry-based approaches are compared here with other state-of-the-art RGB-centric methods showing interest of using polarization imaging.
\keywords{Scene Understanding \and Polarization \and Deep Learning \and Computer Vision}
\end{abstract}

\section{Introduction}
\label{intro}
Joint research in robotics and computer vision is striving to develop approaches achieving machine autonomy. Since there is a great diversity of phenomena in nature, the understanding of urban scenes is very challenging. To reach a form of genericity, the community tends to use massive datasets in combination with deep learning networks. Taking advantage of the abstraction and categorization capabilities of such models, one can learn intermediate representations allowing image understanding. 

Therefore, semantic pixel segmentation and monocular depth estimation contribute to the goal of bringing cognition to robots. 
While one allows the categorization of images into semantic classes at the pixel level, the other estimates distance between sensor and objects. These two applications are crucial since they provide knowledge to the systems and finally define their behavior in front of the surrounding conditions. Traditionally, these techniques tend to allow obstacle recognition and ultimately influence other autonomy factors such as path planning \citep{chen2018importance,ha2017mfnet,zhan2018unsupervised}.

However, the vast majority of approaches rely on feature learning by extracting characteristics from color images. This implies a strong bias since color information is not sensitive to specific phenomena like specularity or transparency. Moreover, the scenes observed in urban environments are prone to this kind of light behavior. For example, when the weather conditions are not optimal (i.e. rain, bright sun, etc.), the color sensing will repeatedly observe saturation or reflections. Thus, through many models designed based on this modality, these aspects are neglected.

Based on this observation, we propose to use the polarization information to define scenes in another space based on physics. Since this information is by definition descriptive of light interaction, we propose two approaches, successively segmentation and depth estimation, relying on these indices to take into account these phenomena. Ultimately, this will allow autonomous robots to be robust to specular behavior.

We therefore propose to use polarization cues and an advantageous image representation to train a convolutional network to obtain accurate segmentation. Addressing the data as the main element of the approach, we proceed to the creation of an augmentation procedure adequate to the physics of the scenes. Without altering the modality properties, we observe notable capabilities to differentiate specular areas from urban scenes. Our evaluation emphasizes the possibility for networks to learn new information and to understand specularity. Thus, we observe better performances when the system observes complex scenes including particular light phenomena.

In a second step, we propose a cost function to infer an accurate depth map from polarization images. Differentiating the non Lambertian areas, we propose to regularize their surface normals using polarimetric features. While this has been a major problem with RGB-centric techniques, we show that the use of this unconventional modality encourages the reconstruction of reflective surfaces. A quantitative evaluation then shows a better robustness to recurrent events in the autonomous vehicle domain such as car reflection and windows specularity. It would then be more feasible to refer to reliable maps that take into account common scenarios, to consider path planning.

In summary, we make the following contributions: (i) a supervised framework for learning segmentation of specular areas through a small amount of polarimetric data; (ii) a novel polarization cues-based loss for depth map estimation; (iii) two monocular specularity invariant approaches to understand urban scenes in realistic conditions.

\section{Related Works}
\label{RelatedWorks}

Humans have an innate ability to understand their surrounding scenes. Despite the presence of specular areas, we are able to overlook their effects. Autonomous systems aspire to the same cognitive abilities and learning-based algorithms tend to increase their perceptive capacities. However, the approaches rely on visual cues that do not constrain some complex problems. Thus, many methods are RGB-centric and propose to perform segmentation or monocular depth estimation based on color cues only. As a consequence, the modality-related inadequacies and flaws affect the different estimation models. Consequently, we propose the alternative of using physics-based vision and specifically polarization sensing.
In the following subsections, we will start by introducing polarization principles, then describe some previous works about scene segmentation and depth estimation.

\subsection{Polarimetry}\label{pola}

\begin{figure}[!h]
    \centering
    \includegraphics[keepaspectratio,width=.8\linewidth]{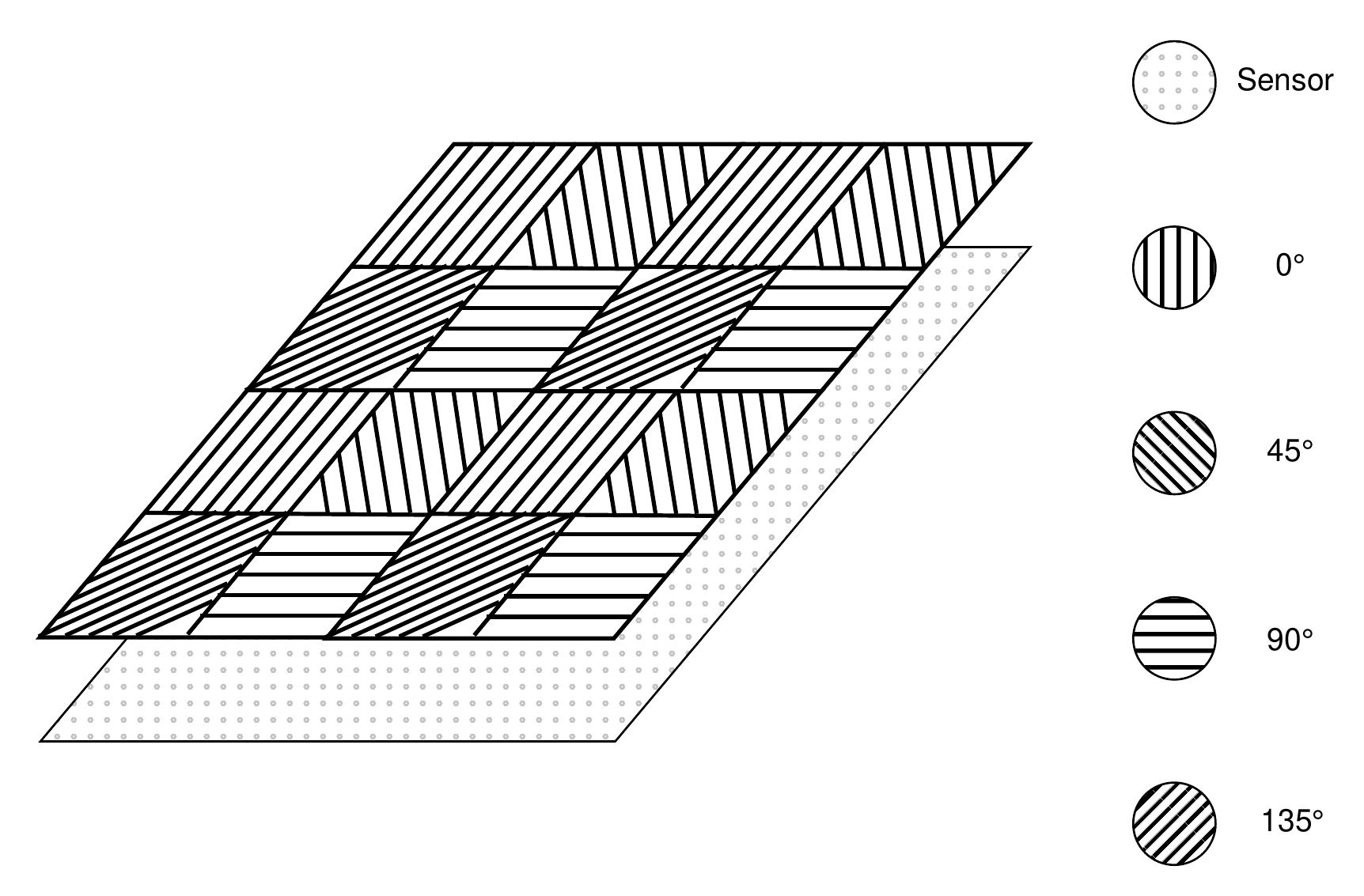}
    \caption{Illustration of a polarimetric camera sensor highlighting the micro-grid with polarizers at different orientations.}
    \label{fig:sensor}
\end{figure}

Polarization \citep{collett2005field} is a non-conventional image space describing the interaction of light with surfaces. Where colorimetry allows to characterize textures, polarimetry acquires the information of reflected waves through polarizers. Indeed, the sensor, shown in Figure \ref{fig:sensor}, is composed of a multitude of 2x2 grid micro-polarizers at different orientations.

\begin{figure}[!h]
    \centering
    \resizebox{.8\linewidth}{!}{
    \colorlet{crystal}{gray}
	
	\def\zangle{-20}
	\def\xangle{20}
	
	\begin{tikzpicture}[x=(\xangle:0.75cm), y=(90:1cm), z=(\zangle:1.5cm),
	>=stealth, line cap=round, line join=round,
	lines/.style={gray!70, thick}, 
	axis/.style={black, thick},
	plate/.style={fill, opacity=0.875},
	markers/.style={darkgray!80, thick},
	markerss/.style={black!30, thick}]

	\begin{scope}[shift={(0,0,3.125)}]
	
	\node [yslant=tan(\zangle), above=0.25cm, align=center,font=\footnotesize] at 
	(1,1,1.5){Linearly Polarized Light};
	
	\begin{scope}[xscale=1.5, yscale=1.5]
	\path [crystal!25, plate] 
	(-1,-1,0) -- (-1,1,0) -- (1,1,0) -- (1,-1,0) -- cycle;
	\path [crystal!50, plate] 
	(-1,-1,0) -- (-1,-1,-0.125) -- (-1,1,-0.125) -- (-1,1, 0) -- cycle;
	\path [crystal!75, plate] 
	(-1,1,0) -- (-1,1,-0.125) -- (1,1,-0.125) -- (1,1, 0) -- cycle;
	\node [yslant=tan(\xangle), text=crystal!50, below, font=\small] at 
	(-1.125,-1,0){Sensor};
	\end{scope}

	\draw [markers] (0,1) -- (0,-1) (-0.5,0) -- (0.5,0);
	
	\draw [axis] (0,0,0) -- (0,0,3);
	
	\foreach \k [evaluate={%
		\i=\k*5.625; \j=\i>0 ? \i-5.625 : 0; 
		\a=90-\i; 
		\b=90-\j; 
		\c=int(mod(\k,4)==0 && sin \a != 0); 
		\d=int(\k+1/4);}] in {0,...,192}{
		\ifodd\d
		\ifnum\c=1
		\draw [->,opacity=0.3] (0,0,\i/360) -- ++(sin \a, sin \a, 0);
		\fi
		\draw [thick, red!80] (sin \a, sin \a, \i/360) -- (sin \b, sin \b, \j/360);
		\else
		\draw [thick, red!80] (sin \a, sin \a, \i/360) -- (sin \b, sin \b, \j/360);
		\ifnum\c=1
		\draw [->,opacity=0.3] (0,0,\i/360) -- ++(sin \a, sin \a, 0);
		\fi
		\fi
	}
	\end{scope}
	
	\begin{scope}[shift={(0,0,6.125)}]
	
	\node [yslant=tan(\zangle), above=0.25cm, align=center,font=\footnotesize] at 
	(1,1,1.5){Unpolarized Light};
	
	\begin{scope}[xscale=1.5, yscale=1.5]
	\path [crystal!25, plate] 
	(-1,-1,0) -- (-1,1,0) -- (1,1,0) -- (1,-1, 0) -- cycle;
	\path [crystal!50, plate] 
	(-1,-1,0) -- (-1,-1,-0.0625) -- (-1,1,-0.0625) -- (-1,1, 0) -- 
	cycle;
	\path [crystal!75, plate] 
	(-1,1,0) -- (-1,1,-0.0625) -- (1,1,-0.0625) -- (1,1, 0) -- cycle;
	\node [yslant=tan(\xangle), text=crystal!50, below, font=\small] at 
	(-1,-1,0){45$^\circ$ Linear Polarizer};
	\end{scope}

	\draw [markerss] (-1.25,0.75) -- (-0.75,1.25);
	
	\draw [markerss] (-1.25,0.25) -- (-0.25,1.25);
	
	\draw [markerss] (-1.25,-0.25) -- (0.25,1.25);
	
	\draw [markerss] (-1.25,-0.75) -- (0.75,1.25);
	\draw [markerss] (-1.25,-1.25) -- (1.25,1.25);
	\draw [markerss] (-0.75,-1.25) -- (1.25,0.75);
	
	\draw [markerss] (-0.25,-1.25) -- (1.25,0.25);
	
	\draw [markerss] (0.25,-1.25) -- (1.25,-0.25);
	
	\draw [markerss] (0.75,-1.25) -- (1.25,-0.75);

	\foreach \k [evaluate={%
		\i=\k*5.625; \j=\i>0 ? \i-5.625 : 0; 
		\a=90-\i; 
		\b=90-\j; 
		\c=int(mod(\k,4)==0 && sin \a != 0); 
		\d=int(\k+1/4);}] in {0,...,192}{
		\ifodd\d
		\ifnum\c=1
		\draw [->,opacity=0.3] (0,0,\i/360) -- ++(sin \a, sin \a, 0);
		\fi
		\draw [thick,red!80] (sin \a, sin \a, \i/360) -- (sin \b, sin \b, \j/360);
		\else
		\draw [thick,red!80] (sin \a, sin \a, \i/360) -- (sin \b, sin \b, \j/360);
		\ifnum\c=1
		\draw [->,opacity=0.3] (0,0,\i/360) -- ++(sin \a, sin \a, 0);
		\fi
		\fi
	}
	
	\foreach \k [evaluate={%
		\i=\k*5.625; \j=\i>0 ? \i-5.625 : 0; 
		\a=90-\i; 
		\b=90-\j; 
		\c=int(mod(\k,4)==0 && cos \a != 0); 
		\d=int(\k+1/4);}] in {0,...,192}{
		\ifodd\d
		\ifnum\c=1
		\draw [->,opacity=0.3] (0,0,\i/360) -- ++(cos \a, 0, 0);
		\fi
		\draw [thick,teal] (cos \a, 0, \i/360) -- (cos \b, 0, \j/360);
		\else
		\draw [thick,teal] (cos \a, 0, \i/360) -- (cos \b, 0, \j/360);
		\ifnum\c=1
		\draw [->,opacity=0.3] (0,0,\i/360) -- ++(cos \a, 0, 0);
		\fi
		\fi
	}
	
	\foreach \k [evaluate={%
		\i=\k*5.625; \j=\i>0 ? \i-5.625 : 0; 
		\a=90-\i; 
		\b=90-\j; 
		\c=int(mod(\k,4)==0 && sin \a != 0); 
		\d=int(\k+1/4);}] in {0,...,192}{
		\ifodd\d
		\ifnum\c=1
		\draw [->,opacity=0.3] (0,0,\i/360) -- ++(0, sin \a, 0);
		\fi
		\draw [thick,orange] (0, sin \a, \i/360) -- (0, sin \b, \j/360);
		\else
		\draw [thick,orange] (0, sin \a, \i/360) -- (0, sin \b, \j/360);
		\ifnum\c=1
		\draw [->,opacity=0.3] (0,0,\i/360) -- ++(0, sin \a, 0);
		\fi
		\fi
	}
	
	\draw [ultra thick, ->] (0,0,3.5) -- (0,0,3);
	
	\end{scope}
	
	\end{tikzpicture}
    }
    \caption{Illustration of a polarizer filtering of unpolarized light.}
    \label{fig:differentiation}
\end{figure}
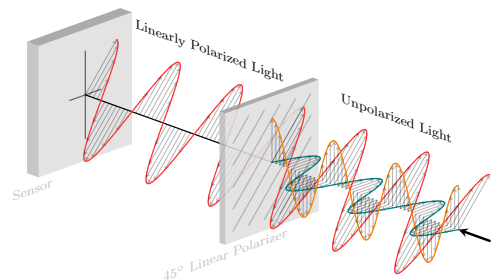

This division of focal plane (DoFP) technology allows the separation of the different components of the received light. Thus, as shown in Figure \ref{fig:differentiation}, the reflected wave is differentiated and consequently, the per-pixel information is influenced by the orientation of the affixed polarizer.

The sensor technology then produces sparse images per orientation. While previously this type of image required thoughtful interpolation \citep{ratliff2009interpolation,li2019demosaicking,zhang2018sparse} that did not alter the image integrity, today's sensors have a permissive resolution resulting in neglecting this sparsity effect. Eventually, from these raw images, one can extract four images reflecting the orientation of each micro-polarizer $P{\{0,45,90,135\}}$. Ultimately, these four components allow to describe the polarization of light through the following Stokes\citep{stokes1851composition} vector:

\begin{equation}
S = \begin{pmatrix}S_0\\S_1\\S_2\\S_3\end{pmatrix} =  \begin{pmatrix}P_0 + P_{90}\\ P_0 - P_{90}\\ P_{45} - P_{135} \\ 0\end{pmatrix}.
\end{equation}

And thanks to a projection in a spherical representation on the Poincaré sphere, one can extract three characteristic images of the polarization:

\begin{equation}
\begin{cases}
\iota = S_0 \\[8pt] 
\alpha = \frac{1}{2} \arctan \frac{S_2}{S_1}\\[8pt] 
\rho =  \frac{\sqrt{S_1^2 + S_2^2}}{S_0}
\end{cases},
\end{equation}

where $\iota$ is the grayscale intensity, $\alpha$ the angle of polarization identifying the orientation of the polarized light with regards to the incident plan and $\rho$ the degree of polarization emphasizing the quantity or the strength of this aforementioned light. Using these different parameters, it is possible to notice the singular link between polarimetric information and surface. This has been defined through Fresnel's equations \citep{fresnel1868oeuvres} defining the relationship between light and material.
Therefore, several authors have taken advantage of this particular information for 3D reconstruction through the Structure from Polarization framework \citep{morel2006active,cui2017polarimetric}. 
Other works tried to improve the methods of perspective reconstruction by inducing polarization indices \citep{berger2017depth}. However, these methods are often effective in constrained environments or with significant prior knowledge. Finally, despite a more differentiable scene description, the use of such a modality remains governed by the constraining Fresnel equations.

However, this relationship implies a strong ability to categorize scenes and differentiate the different surfaces in a scene. While color based techniques only use the scalar value of the incoming light, polarization imaging add the vectorial aspect of it. In addition, while an RGB-centric system will observe saturation and artifacts in front of a specular phenomenon, polarization imaging will better define it. Ultimately, in urban scenes very prone to reflective surfaces, the characterization of such occurrences may benefit the system. 
We therefore propose to evaluate the possibilities of polarization imaging to operate in unconstrained environments, and rely on the remarkable capabilities of deep learning to either learn polarization parameters or abstract Fresnel equations. This is the basis for the use of polarimetry to enable scene understanding tasks such as segmentation and monocular depth estimation.

\subsection{Segmentation}

Semantic segmentation is a predominant domain of computer vision and scene understanding. It consists in assigning a label to every pixel of the image. 
The usual approach consists in providing a massive annotated database and relying on a deep learning network to learn a model infering the semantics from the input image.
Although this methodology has been intensively investigated, the vast majority of the contributions rely on color data sets which implies a textural understanding of the images. Thus, networks such as FCN \citep{long2015fully}, SegNet \citep{badrinarayanan2017segnet}, DilatedNet \citep{yu2015multi} or DeepLab \citep{chen2019towards} have been built to infer a segmentation map. Since the models only learn from the data provided to them, color is the discriminating element in these models. However, the community has tended to increase the complexity of the networks to scale the models and make them more robust. It is seldom unconventional information such as polarization is exploited to constrain the problem upstream.

\begin{figure}[!h]
    \centering
    \includegraphics[keepaspectratio,width=.8\linewidth]{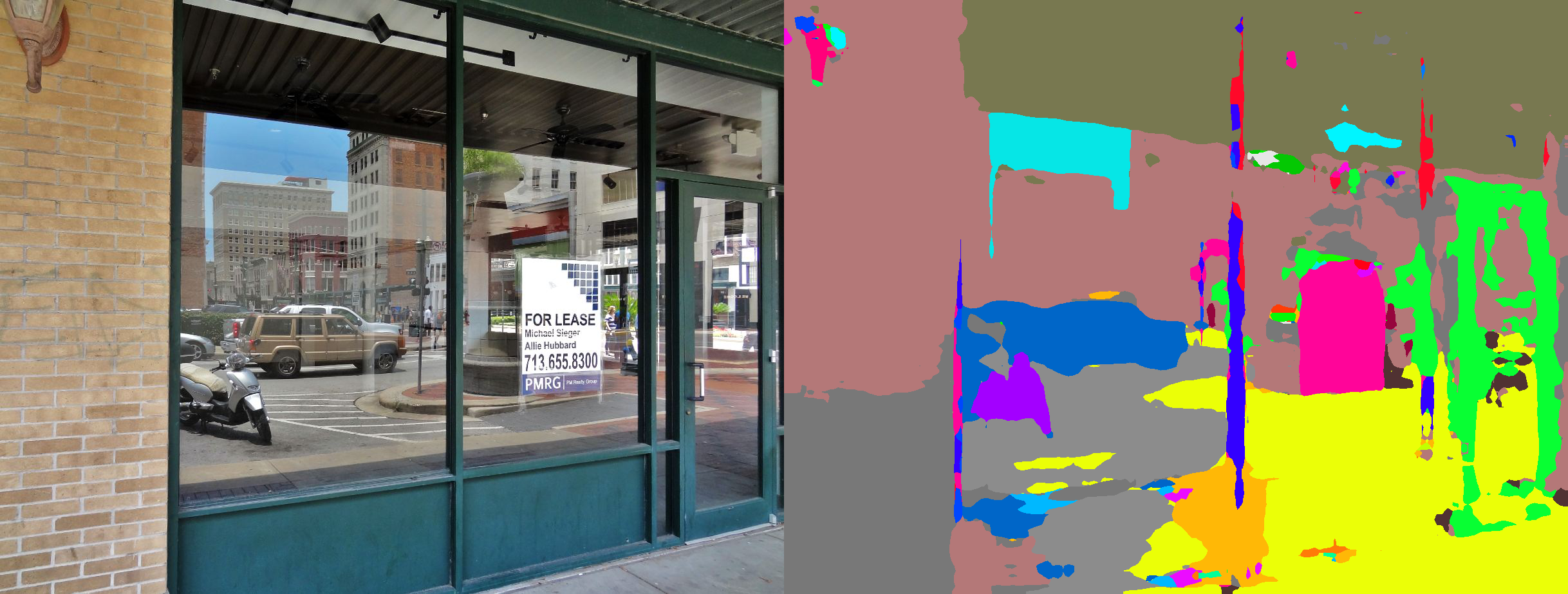}
    \caption{Segmentation results from \citep{zhou2018semantic} showing erroneous estimation on specular surfaces.}
    \label{fig:segrgb}
\end{figure}

As can be seen in the example of Figure \ref{fig:segrgb}, using RGB data can lead to wrong segmentation especially in specular areas. It is therefore important to constrain the problem by using a representation space where object differentiation is facilitated. In this paper, we propose a data-driven segmentation method. Since polarization allows for the physical definition of specular objects, this modality is a potential candidate for more robust feature inference.

\subsection{Depth Estimation}

Depth estimation represents another essential aspect of scene understanding. Defining the distance between the sensor and the object at any point remain a recurrent challenge in computer vision. Such information can ensure the integrity of autonomous systems during their motion by basing decision making on the spacing of obstacles.

Initially, the depth maps were not estimated but rather acquired by means of LiDaR which projects a laser or depth cameras relying on a pattern projection. However, the community has focused on the problem of estimation using an acquisition system with one or more cameras.
Multiple views \citep{favaro2005geometric,parashar2016isometric} or stereo approaches \citep{gennert1988brightness,se2001vision,Woodford2008GlobalPriors} have been used for depth estimation. However, all the stated approaches have drawbacks. The laser-dependent LiDaR is not robust to meteorological elements such as rain and fog or non-Lambertian surfaces. Thus, the robustness of such an acquisition process can be altered by specular surfaces. Projected patterns require favorable conditions and limited distances for optimal observation while image processing based estimation methods can be non-robust and require strong configuration requirements or assumptions such as close distance or limited brightness. 

Over the years, the community has therefore turned to learning-based approaches to alleviate the constraints. While the initial approaches required acquisition to supervise the algorithms \citep{eigen2014depth,liu2015deep,roy2016monocular}, recent methods generalize a model by means of a loss to estimate a disparity map based on physical constraints. Thus, borrowing from the perspective geometry formulation, many networks have demonstrated an ability to estimate dense and robust depth maps in an unsupervised manner.

\begin{figure}[!h]
    \centering
    \includegraphics[keepaspectratio,width=.8\linewidth]{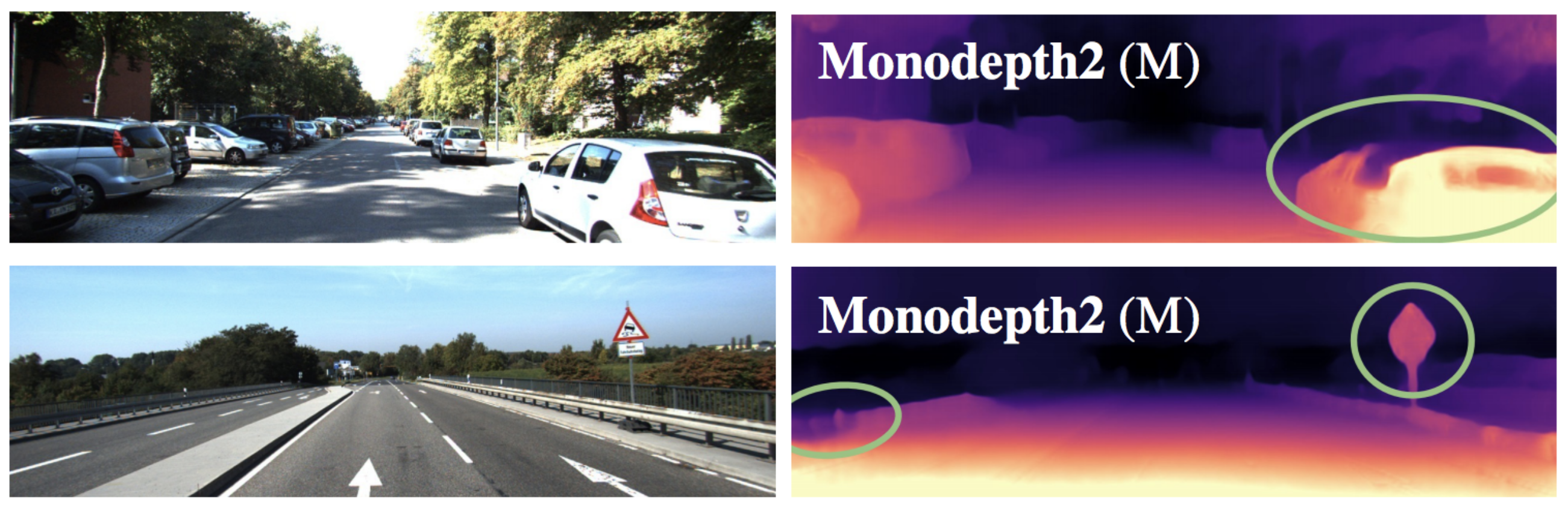}
    \caption{Erroneous depth estimation along specular surfaces. Image borrowed from state-of-the-art method \citep{godard2019digging}.}
    \label{fig:fail}
\end{figure}

Using a generative adversarial network \citep{mehta2018structured}, recurrent online learning \citep{casser2019depth}, 3D motion modeling \citep{luo2019every}, optical flow estimation \citep{ranjan2019competitive} or even finding levers to neglect occlusion \citep{godard2019digging}, a large number of methods have successively shown it is possible to infer a depth map from a single camera system.
This kind of approach is easily exportable to the autonomous vehicle domain since, in addition to being robust, it does not complicate the acquisition system.
On the other hand, most of the networks have been trained on KITTI \citep{Geiger2012CVPR} and only use this dataset and the attached depth as a benchmark. The favorable conditions of these images rarely present surfaces that could be problematic to a sensor not sensitive to physics. Finally, as shown in Figure \ref{fig:fail}, the robustness to specular areas is unproven.

Consequently, we propose using the polarimetric data to consider both diffuse and specular areas. Since this information is sensitive to these occurrences, it would be more adequate to characterize these neglected phenomena. As follows, an inference from polarization could allow autonomous systems to be more robust to such features.

\section{Polarization-based Segmentation}
\label{seg}

Based on the observation that a large number of surfaces present in urban scenes are specular, we proposed to use polarization to characterize these phenomena. In other words, since color imagery does not physically define these occurrences, the concept is to use an image space that facilitates the understanding of these elements.

As stated in Section \ref{pola}, starting from a raw polarimetric image, one can extract four sparse images corresponding to the intensity for each polarizer orientation. The densification being performed by a bilinear interpolation, three characteristic images $\iota$, $\alpha$ and $\rho$ are obtainable through the Stokes vector.
As a result, these three information allow for definition of the polarization states of the objects and therefore define specularity phenomenon. Since deep learning approaches have been shown to be effective in segmenting images, we require a representation to enable polarization encoded images to be used by such architectures. 

\begin{figure}[!h]
    \centering
    \includegraphics[keepaspectratio,width=.8\linewidth]{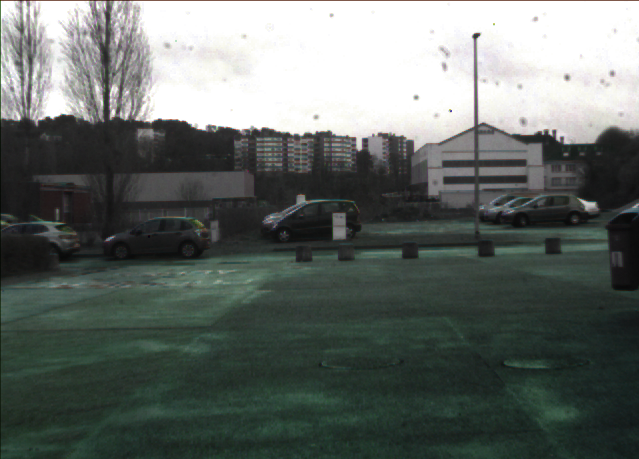}
    \caption{HSL representation of polarization parameters. The ground is colored according to the polarization angle and the intensity of this color is influenced by the degree of polarization.}
    \label{fig:hsl}
\end{figure}

While most of the deep learning based algorithms relies on 3-channel RGB images, this image format does not suit the three parametric information described above. Indeed, while the intensity $\iota$ and the degree of polarization $\rho$ could accommodate such representation, the angle of polarization $\alpha$ represents an orientation and such bounding would reduce the contained information. Different image representations have been proposed for polarimetric imaging but Wolff and Andreou \citep{Wolff1995PolarizationSensors} presented a singular modeling based on the Hue-Saturation-Luminace (HSL) format.
Instead of trying to find an optimal ratio between the components, they chose not to compress or reduce the original bounds of $\iota$, $\alpha$ and $\rho$. The HSL model is composed of three interconnected channels with different limits. Thus, S and L vary in the range $\left[0,100\right]$ while H is a periodic value that varies in the range $\left[0,360\right]$. Interestingly, this model accommodates the polarimetric characteristic images. In addition, such mapping allow for a particular representation, since the hue corresponds to the color shade, the saturation to the strength of the shade and the luminance to the texture or the contrast. Therefore, as shown in Figure \ref{fig:hsl}, this mapping will propose singular behavior since the more polarized is the reflected wave from an area, the more colored is the area. Additionally, the color will allow a direct recognition of the angle thanks to the 360-periodic value.

Finally, despite a particular unnatural coloration of the images and the mapping of polarimetric information on three-channel image, HSL representation is not adapted for network usage. To benefit from color images pre-training and unnatural but representative image formation, one can transform HSL image to RGB image by converting color information into the corresponding RGB values. Consequently, transfer learning can be performed while preserving discriminating properties of the images induced by sensor.

From this image representation, one can train networks to validate the hypothesis that polarization can improve a network's robustness to specularity.

First, an evaluation will be proposed to compare the results using successively color and polarimetric imaging as input. This benchmark being designed to quantify the different cross-modalities, a SegNet network will be used, while disabling the use of advanced contextualization or learning processes.
In a second step, an evaluation will be conducted on the ability of a network to to benefit from the physics of the sensor. Using a state-of-the-art DeepLab v3+ network, we assume that these are the optimal learning conditions. A database can therefore be augmented in multiple manners by taking into account the physics induced by the modality or by neglecting it. 

\subsection{Modality-based evaluation}

\begin{figure}[!h]
    \centering
    \includegraphics[keepaspectratio,width=\linewidth]{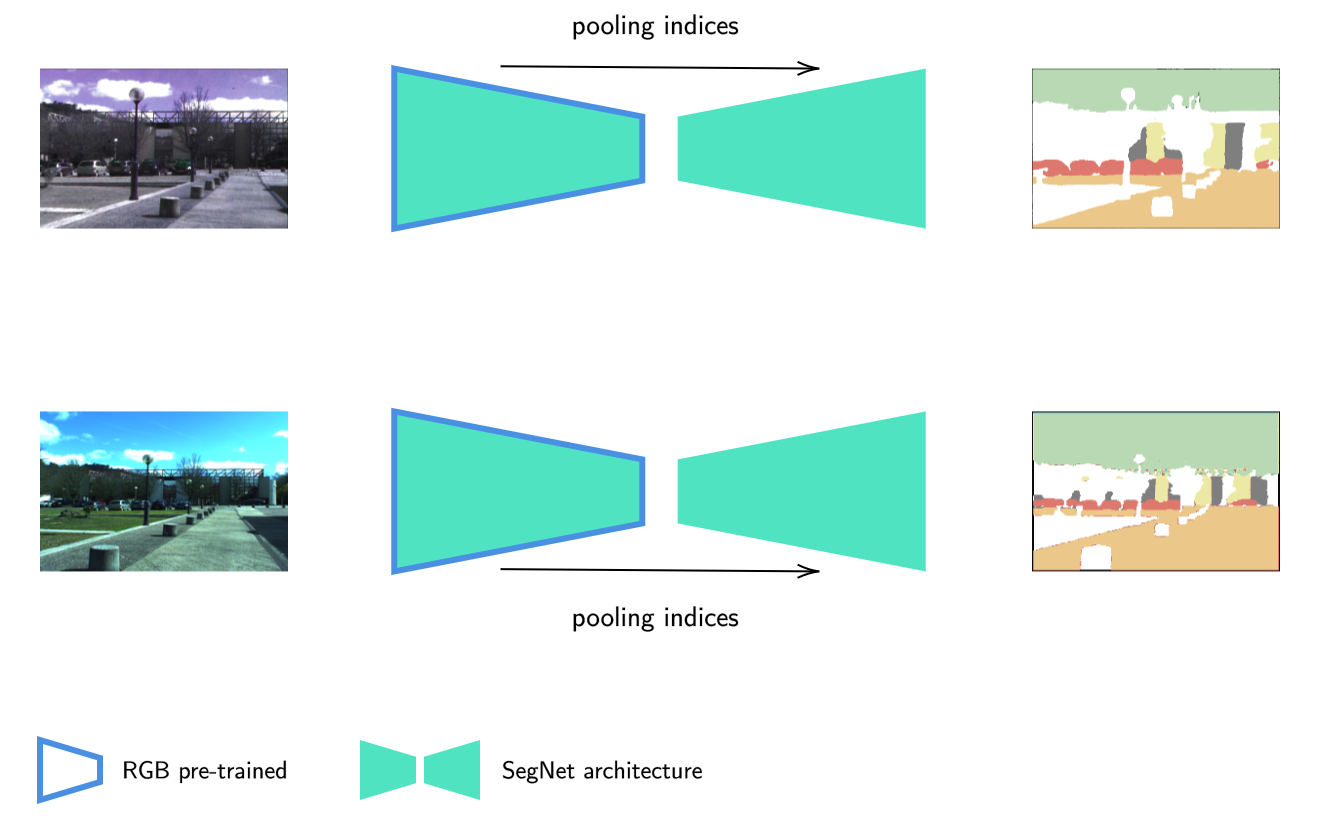}
    \caption{Two pipeline training procedure. Top pipeline receives polarization image and bottom pipeline color image. Both SegNet encoders are initially pre-trained with ImageNet \citep{imagenet_cvpr09} dataset.}
    \label{fig:procseg}
\end{figure}

Using the limited PolaBot\footnote{\url{http://vibot.cnrs.fr/polabot.html}} dataset acquired for comparative learning between modalities, one can train a network for each of the modalities to compare performances on the same scenes \citep{blanchon2019outdoor}.
Furthermore, since SegNet does not embed any attention or contextualization module while being a standard encoder-decoder architecture, this network is the perfect candidate for a fair comparison.
The concept of this segmentation network is mainly based on maxpooling with index to extract the principal components at each activation map. The dimension being successively reduced by the different blocks, at the bottleneck, the initial dimension is recovered through indexed unpooling. 
The spatial information derived from the encoding during the downsampling is thus retranscribed through the indices during the decoding and therefore avoid a naive upsample.
Integrating this concept and assuming that all classes are easily distinguishable, we present the training procedure in Figure \ref{fig:procseg}.

Each of the networks is trained following the same image ordering, which reduces the possible differences in training. The database contains 178 annotated images in both image spaces and includes seven classes that are considered to be of primary importance for the field of autonomous robotics. The models are therefore trained to segment \textcolor{sky}{Sky}, \textcolor{water}{Water}, \textcolor{windows}{Windows}, \textcolor{road}{Road}, \textcolor{car}{Cars}, \textcolor{buildings}{Building} and \textcolor{none}{None} corresponding to other classes.
This two-pipeline procedure allows a quantitative evaluation of the accuracies per class and per modality shown in Table \ref{AccDiff}.

\renewcommand{\arraystretch}{1.5}
\begin{table*}[h]
	\centering
	\caption{Segmentation quantitative results comparing polarimetry and RGB trained models.}
	\label{AccDiff}

	\begin{tabular}{lccccccccc}
         & & \multicolumn{8}{c}{Accuracy} \\
		 & & \textcolor{sky}{@sky} & \textcolor{water}{@water} & \textcolor{windows}{@windows} & \textcolor{road}{@road} & \textcolor{car}{@cars} & \textcolor{buildings}{@building} & \textcolor{none}{@none} & mean \\ \cline{3-10}
		 Model& &  \multicolumn{8}{c}{higher is better}\\ \hline
		Polarimetry & & .753                     & .757                       & \textbf{.828}                        & .778                      & \textbf{.714}                     & \textbf{.876}                          & .789  & \textbf{.785}                     \\ 
		RGB & & \textbf{.895}  & \textbf{.786}  & .445  & \textbf{.784}  & .484  & .678  & \textbf{.834}  & .698                     
	\end{tabular}
	
\end{table*}

Ultimately, polarization influenced network shows increased capabilities to recognize very challenging areas like windows and cars. Considering the dataset, the image quantity being not optimal, the networks most likely have over-fitting issues. Nevertheless, we consider that these results are valid since both approaches were subject to the same constraints. However, this kind of network with limited training data cannot be considered as generic and tends to produce erroneous estimates when the images are different from the ones in training set. 
Overall, polarization based network highlights more robustness which validates the initial hypothesis. While some particular classes are slightly better segmented in color space, one can deduce that the images are advantageous since the color is uniform in the dataset for these specific regions (typically the sky is blue). Following experiment will tackle the task of estimating whether a network is deeply influenced with the physics induced by the modality.

\subsection{Physics-based evaluation}

When comparing the two modalities, polarization demonstrated increased segmentation capabilities of the areas of interest. We subsequently propose an evaluation addressing the physical properties influence  rather than a comparison of intrinsic performance.

\begin{figure}[!h]
    \centering
    \includegraphics[keepaspectratio,width=\linewidth]{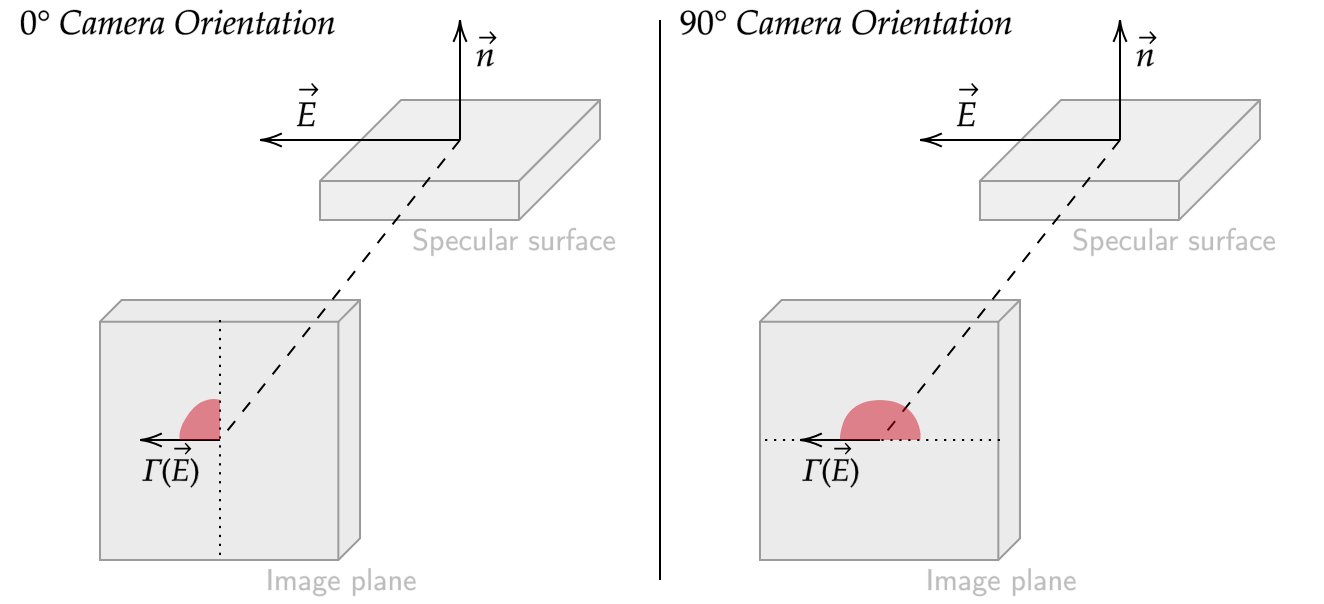}
    \caption{Influence of polarimetric camera rotation on angle of polarization $\alpha$ (in red).}
    \label{fig:augexp}
\end{figure}

In a first step, to have several bases of estimation, one can start from the initial dataset of 178 images and produce two other additional sets using augmention procedures. While one dataset will be subjected to a standard augmentation, the other in addition to the transformation will undergo the regularization necessary to preserve the physical integrity of the images \citep{blanchon2021polarimetric}. As a reminder, polarimetric information is not invariant to pose changes. As shown in Figure \ref{fig:augexp}, a camera rotation, even though the sensor is observing the same surface, will imply a different polarization angle acquisition. This behavior is due to the nature of the polarization angle, which corresponds to the orientation of the electric field $\vec{E}$ with respect to the incident plane. Therefore, the purpose of regularized augmentation is to simulate a physical movement of the camera to create new photorealistic images.

Using three different datasets from the same images, we can evaluate the networks ability to process these data and take advantage of the physical information. We propose to use DeepLab v3+ \citep{chen2019towards} since it is one of the most efficient networks for semantic segmentation. Moreover, this architecture has several essential interests for the task. First, the pooling mechanism conceptualized as ASPP allows an intelligent down-sampling, strongly limiting the effect of discretization of activation maps. This block is also designed to accumulate several receptive fields using atrous convolutions. For semantics, and for our application, this mechanism allows us to include contextual information and thus take advantage of the information contained in more distant neighbor regions compared to standard pooling. 
A second essential point is the use of residual blocks to avoid vanishing gradient problems. This effect is not negligible since it has a direct impact on polarimetric image segmentation. Indeed, despite the hypothesis made in the previous section, one assumption was the classes were distinguishable. This was valid for a simplification of the problem to compare the modalities. However, in real conditions, discrimination by polarization angle is uncertain since a set of $\alpha$ values do not correspond to a specific class. There is thus a more complex connection to define an object. Ultimately, the angle is periodic and depends implicitly on the camera pose. For all these reasons, we rely on a modern architecture to learn any semantic relationships between classes and polarization parameters and thus reveal whether the network is capable of learning from physical polarimetric information.

\begin{figure}[!h]
    \centering
    \includegraphics[keepaspectratio,width=\linewidth]{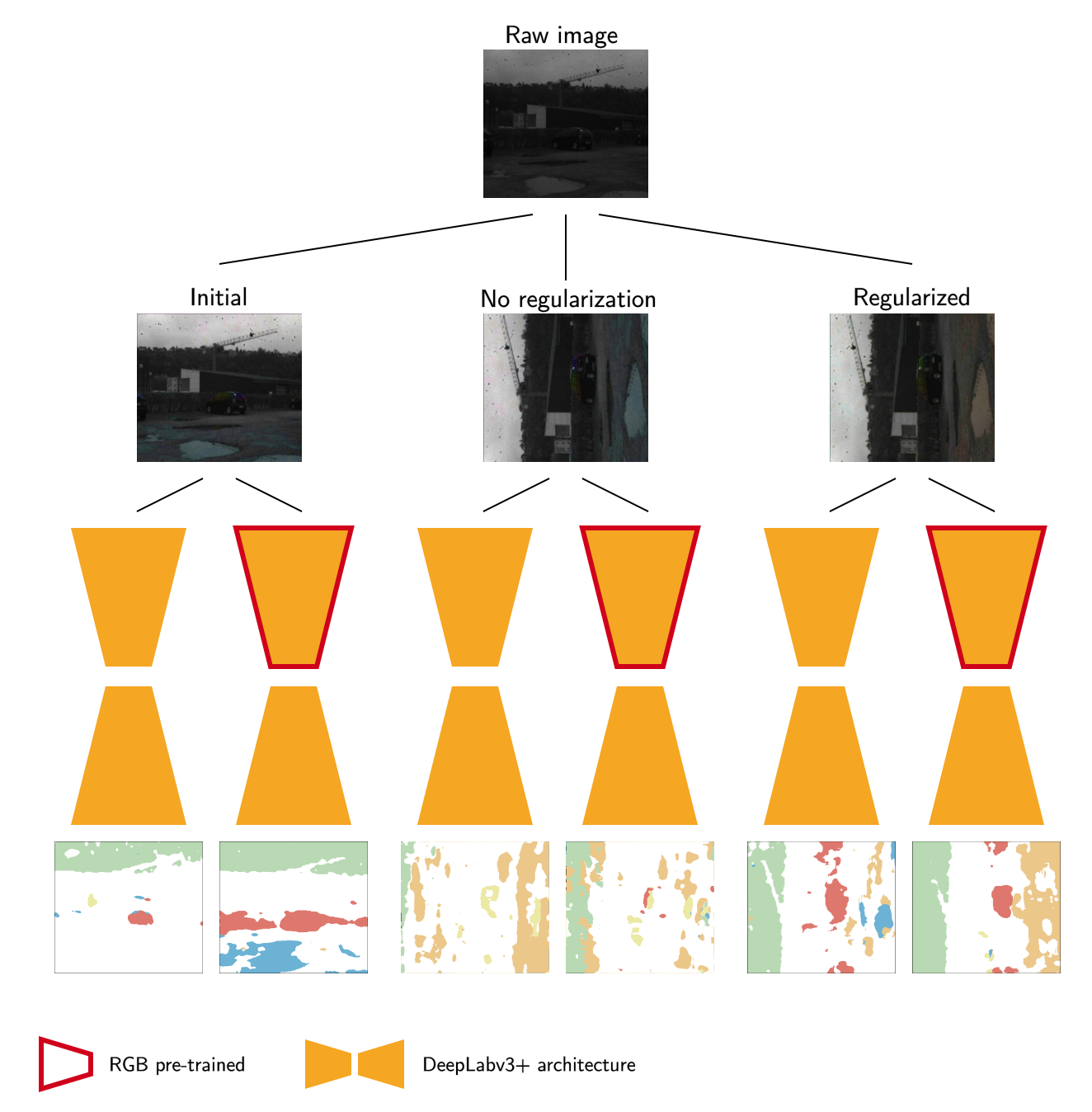}
    \caption{The six different training strategies for physics-infusion evaluation. From raw polarimetric image to physics influenced segmentation through different augmentation techniques.}
    \label{fig:deeplabtrain}
\end{figure}

Finally, the different augmentation strategies are compared using a six-phase training procedure as shown in Figure \ref{fig:deeplabtrain}.
The first network is trained with the original 178 images while the other two are trained with 2136 images obtained through standard or polarization-sensitive regularization respectively.
For testing, an evaluation datastet that does not replicate the type of scenes used for training is used.
The evaluation results are displayed in Table \ref{metricsaug}. We decided to extract metrics by specific classes since they are of interest for scene understanding applications in urban context. Consequently, we highlight the classes cars, windows and water since they are traditionally difficult to segment using RGB-centric methods.

\renewcommand{\arraystretch}{1.5}
\begin{table*}[ht]
\centering
\caption{Quantitative evaluation of Deeplab v3+ specular surfaces segmentations with respect to augmentation procedure and pre-training. Metrics computed excluding building class are denoted $\backslash B$. Mean metrics are computed on all the seven classes while Mean $\backslash B$ excludes the building class.}
\label{metricsaug}
\resizebox{\textwidth}{!}{%
\begin{tabular}{cclccccclccccc}
 &
   &
   &
  \multicolumn{5}{c}{IoU} &
   &
  \multicolumn{5}{c}{Recall} \\
 &
   &
   &
  \textcolor{water}{@water} &
  \textcolor{windows}{@windows} &
  \textcolor{car}{@cars} &
  Mean &
  Mean $\backslash B$ &
   &
  \textcolor{water}{@water} &
  \textcolor{windows}{@windows} &
  \textcolor{car}{@cars} &
  Mean &
  Mean $\backslash B$ \\ \cline{4-8} \cline{10-14} 
\multicolumn{1}{l}{Augmentation} &
  \multicolumn{1}{l}{PreTraining} &
   &
  \multicolumn{5}{c}{higher is better} &
   &
  \multicolumn{5}{c}{higher is better} \\ \hline
\multirow{2}{*}{None} &
  No &
   &
  .40 &
  .20 &
  .20 &
  .30 &
  .32 &
   &
  .35 &
  .15 &
  .22 &
  \textbf{.50} &
  \textbf{.50} \\
 &
  Yes &
   &
  .54 &
  .10 &
  .43 &
  .33 &
  .34 &
   &
  \textbf{.42} &
  .15 &
  .57 &
  .43 &
  \textbf{.50} \\ \hline
\multirow{2}{*}{Not regularized} &
  No &
   &
  .001 &
  .03 &
  .12 &
  .14 &
  .13 &
   &
  .35 &
  .25 &
  .15 &
  .31 &
  .28 \\
 &
  Yes &
   &
  .10 &
  .03 &
  .19 &
  .21 &
  .20 &
   &
  .35 &
  .22 &
  .23 &
  .37 &
  .33 \\ \hline
\multirow{2}{*}{Regularized} &
  No &
   &
  .63 &
  .13 &
  .46 &
  \textbf{.43} &
  \textbf{.50} &
   &
  .39 &
  .21 &
  \textbf{.60} &
  .43 &
  \textbf{.50} \\
 &
  Yes &
   &
  \textbf{.70} &
  \textbf{.26} &
  \textbf{.47} &
  .37 &
  .38 &
   &
  .35 &
  \textbf{.26} &
  .48 &
  .42 &
  .38
\end{tabular}%
}
\end{table*}

\subsection{Discussion}

This step-by-step experimentation allowed us to evaluate the hypothesis that polarization would provide better capability to characterize urban scenes and especially specular surfaces.
First, a fair comparison between color and polarimetric imagery quantified the advantage of polarization in addressing such areas. Despite the limited amount of data, it is clear that this new information can be beneficial to a deep learning network.
However, it was impossible to validate if the model had been influenced with the physical properties of the image. In response, we proposed to evaluate this capability by training six models observing data whose physics had been altered or not. This implementation highlighted the ability of the model to learn polarimetric information. Thus, this experimentation emphasizes it is better not to augment than to ignore the properties of the modality during the augmentation process..

In this section, we proposed an end-to-end pipeline for semantic segmentation of urban scenes using polarization that allows the detection of  specular areas which are mostly hazardous in such environments and can affect the integrity of autonomous systems.
Therefore, taking into account specularity is essential to obtain robust recognition algorithms.

\section{Towards Polarimetric Monodepth}
\label{depth}

A second axis of this urban scene understanding framework using polarization is depth estimation. However, contrary to the segmention method described in Section \ref{seg} which is supervised, we adapt here an unsupervised approach and use the physics of the modality to provide additional constraint terms enabling accurate scene reconstruction. Since polarimetry characterizes light interaction and specularity effects, a model can be driven by the polarization state of the light \citep{blanchon2021p2d}.

Besides the deep learning approach, Berger et al. \citep{berger2017depth} have investigated the possibility of constraining the stereo reconstruction problem using polarization priors. Therefore, by using an adapted calibration allowing the direct mapping between polarization angle and orientation of the normal, they demonstrates it is possible to reconstruct highly specular surfaces. Although the process adds a rectification term for these areas, the acquisition process is too restrictive to be operable in real conditions. For this reason, we propose several terms to impact the depth estimation of reflective objects operable in real-world condition. We take advantage of the abstraction capabilities of deep learning model networks to neglect the consideration of Fresnel equations governing relation between light and materials.

\begin{figure}[!h]
    \centering
    \includegraphics[keepaspectratio,width=\linewidth]{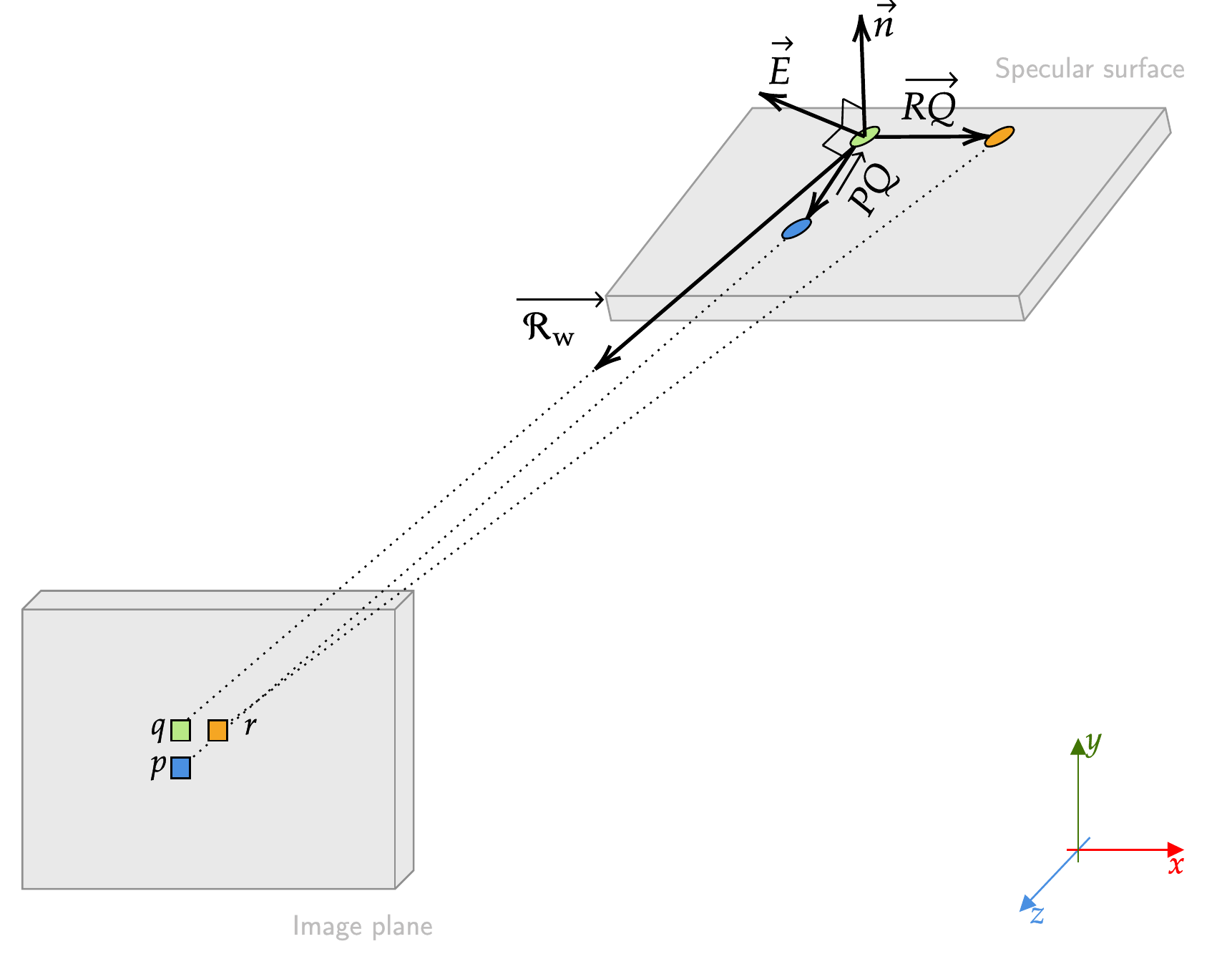}
    \caption{Electric field estimation from three neighbor points in 2D image plane.}
    \label{fig:exppola}
\end{figure}

To constitute a new penalty term representing the difference between polarization angle and orientation of the normal, one can define the relation that links them such that:

\begin{equation}
    \alpha = tan^{-1} {\Gamma(\vec{E})},
\end{equation}

with $\vec{E}$ the electric field defined as a perpendicular vector with regard to the normal for specular surfaces and $\Gamma()$ the image plane reprojection operation. As shown in Figure \ref{fig:exppola}, from three 3D-projected point belonging to the same surface, one can deduce the surface normal.

Ultimately, the electric field represents the vector perpendicular to both $\vec{\mathcal{R}_w}$ and the surface normal $\vec{n}$. Considering two formulations, it is possible to define $\vec{E}$ angle with regard to incident plane according to:

\begin{equation}\label{a}
    \Gamma(\vec{E}_r) = \Gamma\Big( \big(\vec{PQ} \times \vec{RQ} \big) \times \vec{\mathcal{R}_w}\Big),
\end{equation}

or, if we neglect the impact of the reflected wave and consider each points to be aligned with the optical center:

\begin{equation}\label{b}
    \Gamma(\vec{E}_a) = \Gamma\Big(\big(\vec{PQ} \times \vec{RQ} \big) \Big) \pm \frac{\pi}{2} \quad \mod\pi.
\end{equation}

\begin{figure}[!h]
    \centering
    \includegraphics[keepaspectratio,width=\linewidth]{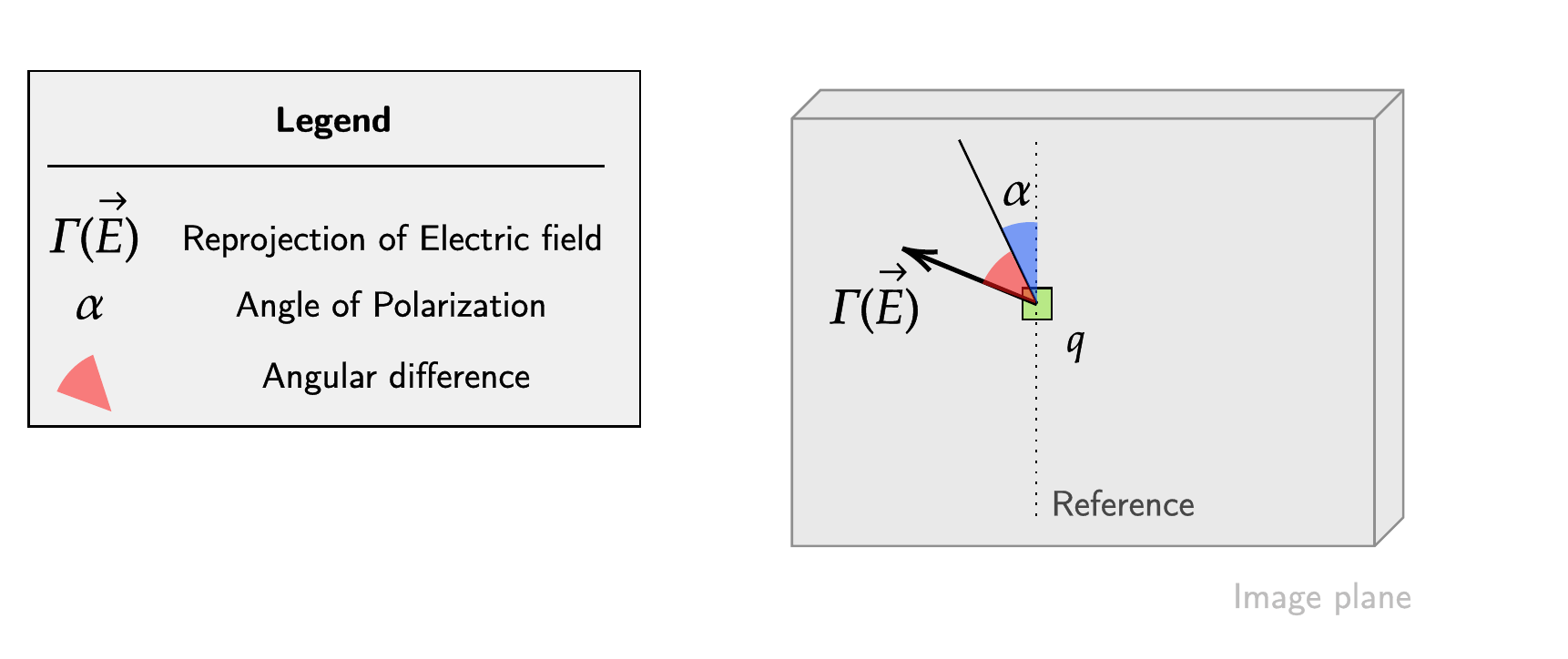}
    \caption{Representation of angular error on image plane.}
    \label{fig:angerr}
\end{figure}

Note, $E_a$ and $E_r$ are respectively the \textit{approximated} or the \textit{real} electric field. Consequently, projecting this obtained vector onto the image plane, one can deduce a bounded angular error $\mathcal{A}$ (represented in Figure \ref{fig:angerr}) such that:

\begin{equation}
    \mathcal{A} = \Big| tan \Big( tan^{-1} {\Gamma(\vec{E})} - \alpha \Big) \Big|.
\end{equation}

Using absolute tangent allow shifting from an angular radian error into a $[0,\infty[$ interval error. In addition, $\vec{E}$ can be estimated following either equation \ref{a} or equation \ref{b} and consequently imply $\mathcal{A}$ to be either \textit{exact} or \textit{approximated}. Because of the polarization properties, this formulation is valid iff the observed surface is specular. Since we do not observe any specularity index, we decide to use the degree of polarization to define the nature of the objects. Therefore, one can define an empirical threshold of 0.4 deemed sufficient descriptor to validate the formulation. Additionally, to quantify and scale the error, we propose to use this truncated polarization degree to complete this angle difference penalty term as follows: 

\begin{equation}
    L_p = \rho \mathcal{A}_r,
\end{equation}

or when using electric field orientation approximation:

\begin{equation}
    L_p = \rho  \min( | \mathcal{A}^{+\frac{\pi}{2}}_a |, | \mathcal{A}^{-\frac{\pi}{2}}_a |).
\end{equation}

With $\rho$ the degree of polarization, this regularization term allows polarization to be taken into account but it remains to comply with standard methods including reprojection and smoothness.
The error based on the perspective geometry while taking into consideration the occlusion are those borrowed from \citep{godard2019digging}:

\begin{equation}
	L_r = \min_{t^\prime} \: pe(I_t, I_{t^\prime \rightarrow t}),
\end{equation}

with $I_{t^\prime \rightarrow t}$ the reprojected second image onto the source image plane and $pe$ the reconstruction error:

\begin{equation}
pe(I_a, I_b) = \frac{\beta}{2} (1- SSIM(I_a, I_b)) + (1-\beta) ||I_a - I_b||_1.
\end{equation}

This term consider both the photometric error by considering dissimilarity measure and L1 norm while the $min$ reduces image boundaries artifacts.
Finally, smoothness term consists in densifying disparity map, smoothing surfaces and preserving salient edges. Contrarily to Godard et al., we propose using the second order smoothness which validity has been proven in \citep{Woodford2008GlobalPriors}. Embedding edge preserving term, we define:

\begin{equation}
	L_s = |\delta^2_xd_t^*| e^{-|\delta^2_x I_t|} + |\delta^2_yd_t^*| e^{-|\delta^2_y I_t|},
\end{equation}

with $d_t^* = d_t / \bar{d_t}$ the mean-normalized inverse depth and $\delta^2$ the second order derivative. Accumulating the three term, the general loss can be defined as:

\begin{equation}
    \Lambda = \beta (\mu L_r) + \lambda L_s + \gamma L_p,
\end{equation}

with $\mu$ the binary mask considering occlusion and object stationary behavior and $\beta$, $\lambda$ and $\gamma$ three empirical scaling factors managing the impact of each term onto the final reconstruction.

\begin{figure}[!h]
    \centering
    \includegraphics[keepaspectratio,width=\linewidth]{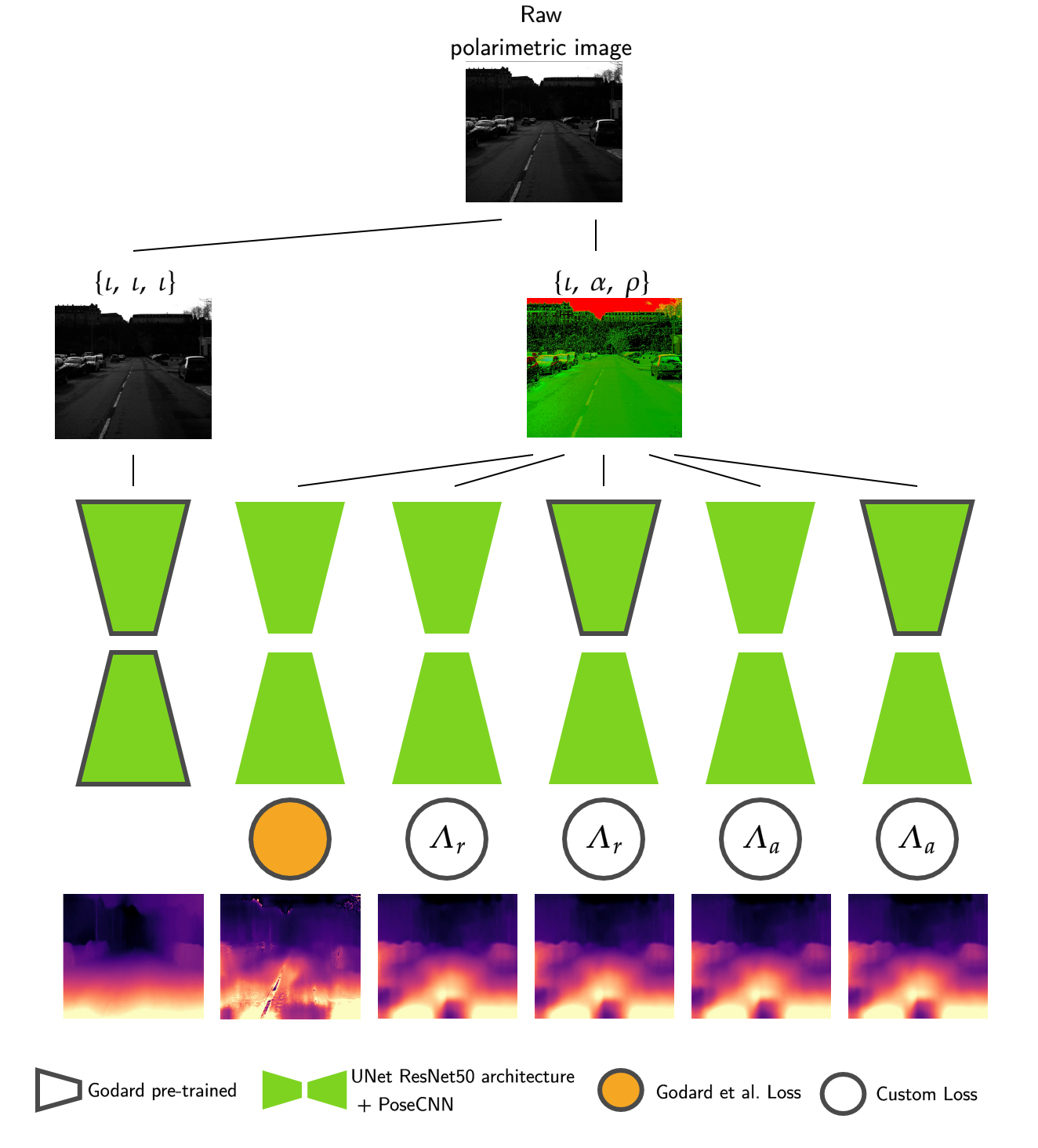}
    \caption{Training procedure for evaluation. Every trained network embed pose estimation through PoseCNN. The indices $_r$ and $_e$ refers respectively to \textit{real} and \textit{approximate} of the electrical field.}
    \label{fig:traindepth}
\end{figure}

$\Lambda$ being defined, we define the training framework allowing a comparison between networks trained with the different proposed losses and the state-of-the-art method. As shown in Figure \ref{fig:traindepth}, multiple network have been trained to compare our methods.

We choose to evaluate the initial network of Godard et al. with a concatenation of polarimetric intensity images $\{\iota, \iota, \iota\}$ and to train 5 networks, three of which are end-to-end with polarization parameters concatenation $\{\iota, \alpha, \rho\}$. Since it has been validated network are able to take advantage of transfer learning, we decided to evaluation two networks while pre-training the ResNet 50 encoder with \citep{godard2019digging}.
Ultimately, Table \ref{tab:resdepth} display quantitative results for each methods.

\renewcommand{\arraystretch}{1.5}
\begin{table*}[ht]
\centering
\caption{Quantitative evaluation of different depth estimation networks. G\_$\{\iota,\iota,\iota\}$ is network proposed in \citep{godard2019digging} and P2D\_G stands for end-to-end training using Godard et al. loss, hence, no polarimetric regularization. In addition $^p$ denotes pre-trained and  $_a$ and $_r$ represent respectively \textit{approximated} and \textit{real} electric field estimation.}
\label{tab:resdepth}
\resizebox{\textwidth}{!}{%
\begin{tabular}{lllcccclccc}
 &
   &
   &
  Abs Rel &
  Sq Rel &
  RMSE &
  RMSE log &
   &
  $\delta > 1.25$ &
  $\delta > 1.25^2$ &
  $\delta > 1.25^3$ \\ \cline{4-7} \cline{9-11} 
Type &
  Network &
   &
  \multicolumn{4}{c}{lower is better} &
   &
  \multicolumn{3}{c}{higher is better} \\ \hline
\multirow{6}{*}{Raw} &
  G\_$\{\iota,\iota,\iota\}$ &
   &
  0.471 &
  10.809 &
  25.161 &
  0.680 &
   &
  \textbf{0.485} &
  0.707 &
  0.804 \\
 &
  P2D\_G &
   &
  0.482 &
  9.144 &
  22.332 &
  0.617 &
   &
  0.431 &
  0.695 &
  0.838 \\
 &
  P2D\_$^p\Lambda_a$ &
   &
  0.512 &
  4.930 &
  19.611 &
  0.701 &
   &
  0.358 &
  0.611 &
  0.755 \\
 &
  P2D\_$\Lambda_a$ &
   &
  0.555 &
  5.267 &
  18.447 &
  0.643 &
   &
  0.363 &
  0.619 &
  0.776 \\
 &
  P2D\_$^p\Lambda_r$ &
   &
  0.416 &
  4.920 &
  20.354 &
  0.977 &
   &
  0.401 &
  0.668 &
  0.816 \\
 &
  P2D\_$\Lambda_r$ &
   &
  \textbf{0.355} &
  \textbf{3.857} &
  \textbf{17.657} &
  \textbf{0.408} &
  \textbf{} &
  \textbf{0.485} &
  \textbf{0.779} &
  \textbf{0.888} \\ \hline
\multirow{6}{*}{Cropped} &
  G\_$\{\iota,\iota,\iota\}$ &
   &
  0.533 &
  14.050 &
  29.312 &
  0.780 &
   &
  0.449 &
  0.658 &
  0.771 \\
 &
  P2D\_G &
   &
  0.415 &
  11.247 &
  25.899 &
  0.678 &
   &
  0.467 &
  0.729 &
  0.850 \\
 &
  P2D\_$^p\Lambda_a$ &
   &
  0.413 &
  6.054 &
  22.360 &
  0.654 &
   &
  0.403 &
  0.684 &
  0.836 \\
 &
  P2D\_$\Lambda_a$ &
   &
  0.323 &
  5.706 &
  21.040 &
  0.508 &
   &
  0.481 &
  0.767 &
  0.865 \\
 &
  P2D\_$^p\Lambda_r$ &
   &
  0.403 &
  6.018 &
  23.516 &
  0.766 &
   &
  0.362 &
  0.629 &
  0.786 \\
 &
  P2D\_$\Lambda_r$ &
   &
  \textbf{0.280} &
  \textbf{4.650} &
  \textbf{20.441} &
  \textbf{0.475} &
  \textbf{} &
  \textbf{0.524} &
  \textbf{0.794} &
  \textbf{0.911} \\ \hline
\multirow{6}{*}{Specular} &
  G\_$\{\iota,\iota,\iota\}$ &
   &
  0.341 &
  8.249 &
  7.236 &
  0.306 &
   &
  0.666 &
  0.808 &
  0.896 \\
 &
  P2D\_G &
   &
  0.208 &
  2.248 &
  5.491 &
  0.233 &
   &
  0.639 &
  0.877 &
  0.952 \\
 &
  P2D\_$^p\Lambda_a$ &
   &
  0.210 &
  0.434 &
  1.071 &
  0.271 &
   &
  0.714 &
  0.867 &
  0.952 \\
 &
  P2D\_$\Lambda_a$ &
   &
  0.189 &
  0.542 &
  1.348 &
  0.231 &
   &
  0.768 &
  0.914 &
  0.951 \\
 &
  P2D\_$^p\Lambda_r$ &
   &
  0.207 &
  0.402 &
  1.290 &
  0.314 &
   &
  0.675 &
  0.864 &
  0.924 \\
 &
  P2D\_$\Lambda_r$ &
   &
  \textbf{0.146} &
  \textbf{0.231} &
  \textbf{0.897} &
  \textbf{0.193} &
  \textbf{} &
  \textbf{0.760} &
  \textbf{0.941} &
  \textbf{0.983}
\end{tabular}%
}
\end{table*}

Polarization image allows to infer directly on the normal to the planes. Constraining this problem by considering only the specular surfaces, an angular error can be formulated as a minimizable term of the overall reconstruction loss.
We have proposed several methods to evaluate the performance based on a pre-training and/or an approximate electric field estimate. Realistically, the consideration of the reflected wave $\vec{R_w}$ is complex and involves computations in 3D space leading to the accumulation of uncertainties. However, the strategy considering that the electric field $\vec{E}$ is always perpendicular to both reflected wave and the normal for specular surfaces does not allow to observe the same robustness despite the simplicity of this formulation.
We also proposed to evaluate our methods in competition with the original method of Godard et al. \citep{godard2019digging} by using only an intensity concatenation to mimic the RGB. This approach allows a reliable comparison through pixel alignment since $\iota$ comes from the acquisition but one can consider the network is handicapped by the absence of color. Ultimately, we proposed to train a network with the concatenation of the polarimetric information $\{\iota, \alpha, \rho\}$ while neglecting the polarimetric term of the loss to evaluate whether the other methods were valid or $L_p$ term did not contaminate the network. 

To assess the methods, we propose to use three strategies consisting in comparing the whole images, the cropped images and only the specular areas. We justify cropping since the maps deduced from polarization tend to produce short range artifacts. We assume that this is due to two factors, the proximity of the pixels implies drastic changes in polarization angles and the polarimetric information is not invariant to pose. Since the acquisition conditions of the test dataset are not identical to the training dataset, then this phenomenon can be explained, but it is indicative of a lack of genericity of the model.

To conclude briefly, P2D\_$\Lambda_r$ without pre-training and with a loss dependent on $\vec{R_w}$ gives the best results and proves that, firstly, the network can be infused with the physics of imaging, and secondly, the network struggles to change domain when using this kind of loss.

\section{General Discussion}
\label{gedisc}

We proposed a combination of scene understanding approaches from a single raw polarimetric image. To demonstrate the contribution of polarization indices to the definition of urban scenes, we first proposed a naive approach to segmentation involving polarization. This initial method emphasized the usefulness of an unconventional modality sensitive to the polarization state of the light, especially for segmenting specular areas. In a second step, we closed the segmentation line by verifying the ability of the networks to learn physical properties. Using augmentation, this highlighted that the networks could benefit from this information, but also that the augmentation had to be in agreement with the modality.
Building on the conclusions drawn from the segmentation approaches, we proposed a depth estimation pipeline with a single view. This time, since it is difficult to acquire robust ground truth depth maps, we employed a self-supervised approach. Constraining the angle divergence of the normals to the specular surfaces, we demonstrated that we could use polarization to obtain an accurate reconstruction. With the aim of overcoming color-based methods weaknesses, we show depth estimation on specular surfaces is improved.

\begin{figure}[!h]
    \centering
    \includegraphics[keepaspectratio,width=\linewidth]{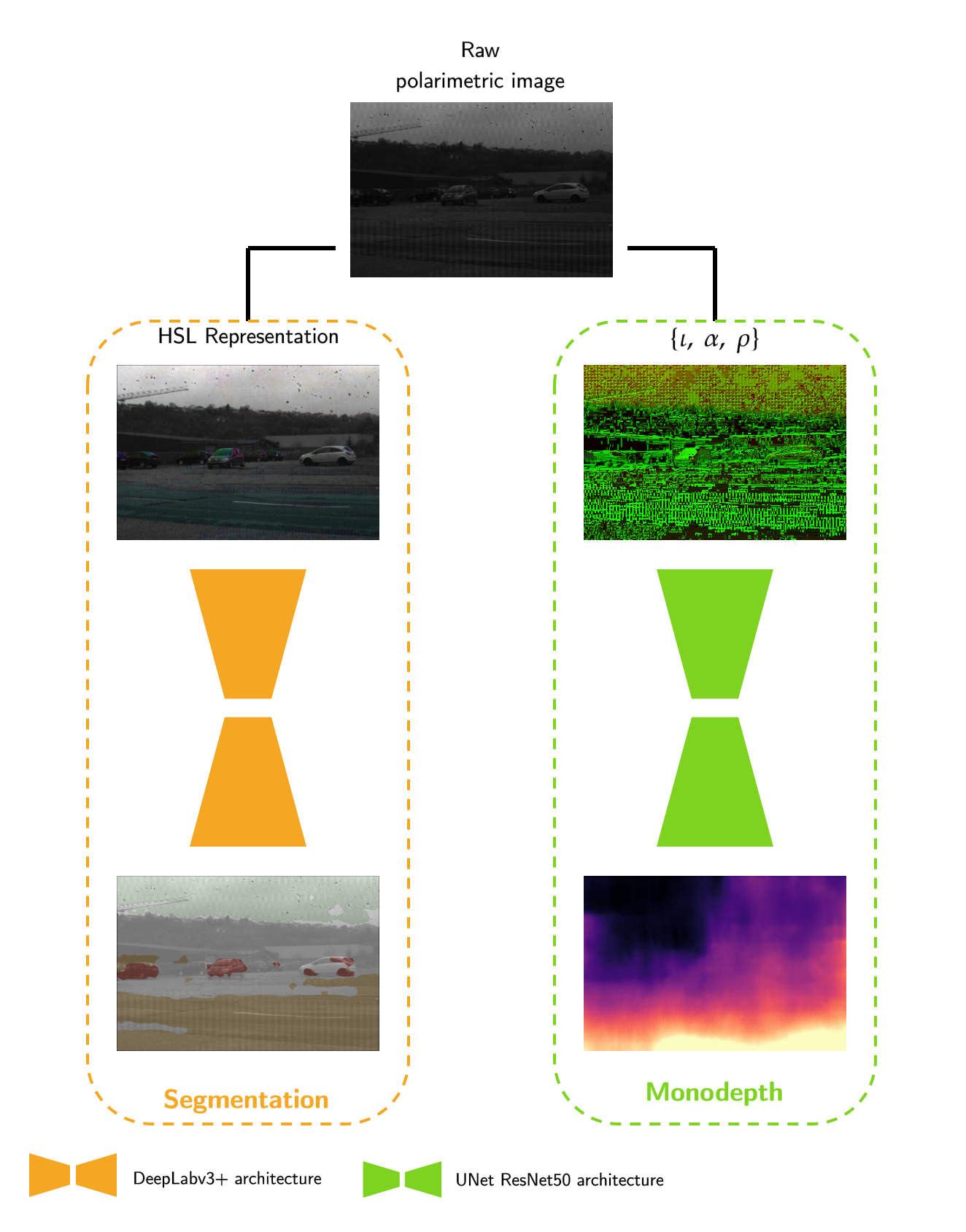}
    \caption{Two-sided scene understanding pipeline through polarization cues.}
    \label{fig:doublepipe}
\end{figure}

\begin{figure*}[!h]
    \centering
    \includegraphics[keepaspectratio,width=.95\linewidth]{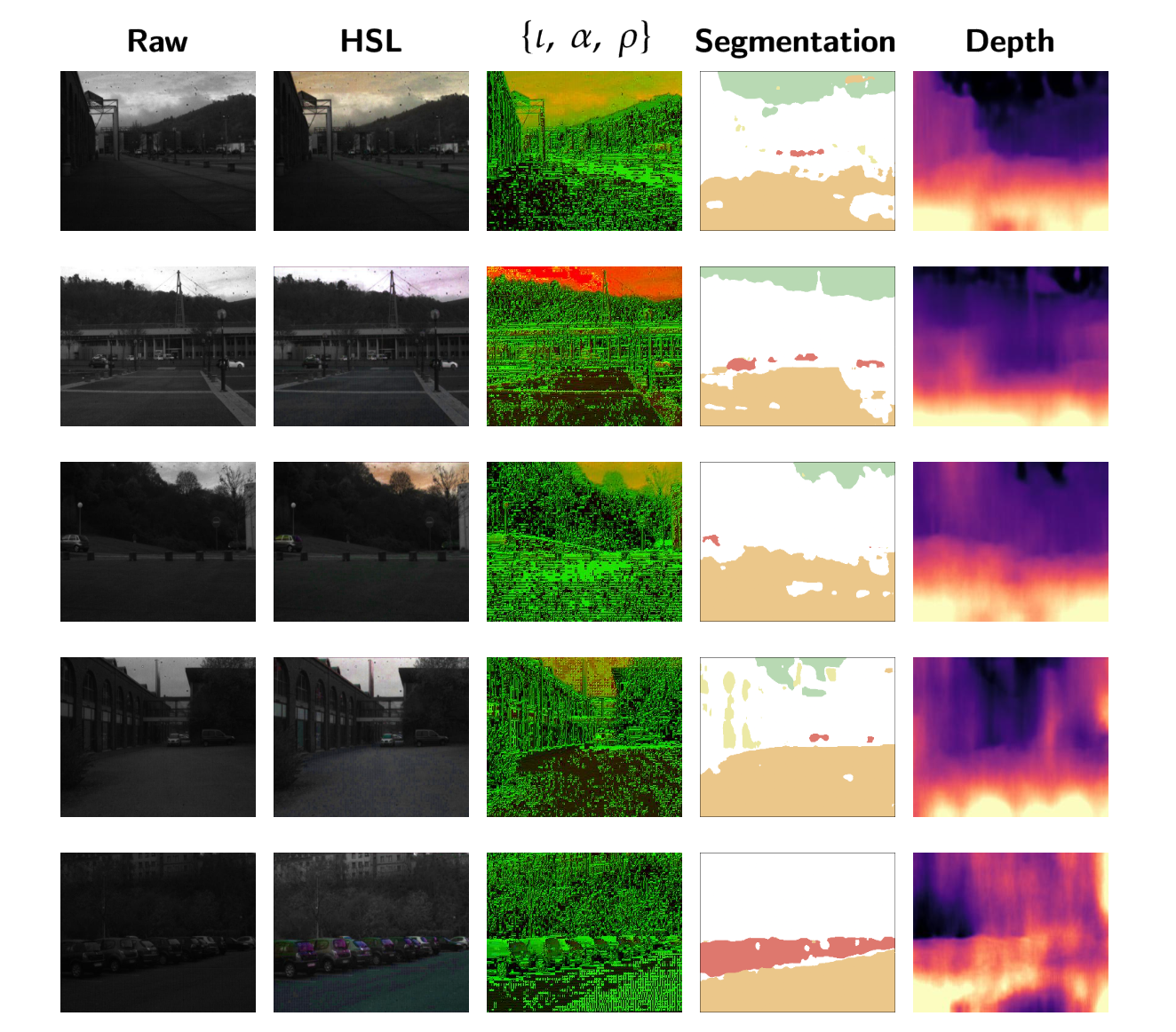}
    \caption{Qualitative evaluation of two-sided scene understanding pipeline using polarization.}
    \label{fig:quant}
\end{figure*}

Finally, these different approaches allow us to develop a double channel method of scene understanding from a polarimetric monocular image shown in Figure \ref{fig:doublepipe}. Ultimately, a qualitative evaluation is shown in Figure \ref{fig:quant}.

\section{Conclusion}
\label{conclu}

In this paper, we proposed a first step towards a robust characterization of urban scenes through polarization. Constrained by the amount of information available, we have developed both segmentation and monodepth approaches to show the interest of including polarization cues into these algorithms. It is notable that the data is critical in any domain but specifically in unconventional imaging since it is scarce. The methods using this kind of information are then at a disadvantage and tend to be neglected whereas they have a more interesting description capability. We have proposed a first segmentation pipeline that defines specular surfaces in a more optimal way. Of course, the robustness of such a network is not proven due to the dataset, but, as a proof of concept, this allowed us to show that
deep learning based methods can benefit from polarization for a more robust urban scene characterization. 

In a second step, we proposed a loss regularization term to refine the depth maps from the so-called monodepth networks. Taking advantage of the polarization indices, we have defined several functions and approaches to evaluate the validity of this data for the task. 
Therefore, we have shown an increased robustness for the estimation of urban scenes and specifically for specular surfaces. However, it is notable that the estimates can be erroneous and thus the method is not fully robust.
As a proof of concept, we have tried to improve the process in a joint training manner. This approach greatly facilitates the training process but tends to negatively impact the estimation process using only photometric loss. This is due to the accumulation of terms that may conflict and impact the overall quality of the network training. One technique would be to use modern fusion processes to improve the depth maps without altering the prior RGB-centric network estimates. 

Finally, this paper proposes an initial step towards urban scenes understanding using polarization. The different experiments have proven the interest of this physics-based modality to characterize complex scenes, especially reflective areas.

A major advance would be the acquisition of a large annotated polarimetric dataset, which would allow for a deeper analysis and a fair and viable comparison.

\begin{acknowledgements}
This work was supported by the French National Research Agency through ANR ICUB (ANR-17-CE22-0011). We gratefully acknowledge the support of NVIDIA Corporation with the donation of GPUs used for this research. 
\end{acknowledgements}

%
\section*{Conflict of interest}

The authors declare that they have no conflict of interest.

\printbibliography

  \begin{wrapfigure}{l}{25mm} 
    \includegraphics[width=1in,height=1.25in,clip,keepaspectratio]{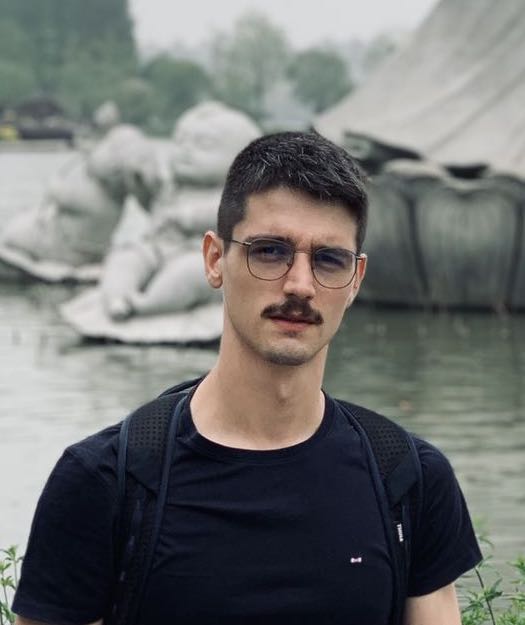}
  \end{wrapfigure}
  \textbf{Marc Blanchon} is a third-year Ph.D Student in Universit\'e Bourgogne Franche-Comt\'e in the EMR VIBOT CNRS 6000 team. He recieved a Bachelor's degree and a Master's degree both in Computer Vision from Universit\'e Bourgogne Franche-Comt\'e in 2016 and 2018 respectively. His main research interests are computer vision for autonomous  vehicles and physics-based deep learning. \vspace{.1cm}\\
  
  \begin{wrapfigure}{l}{25mm} 
    \includegraphics[width=1in,height=1.25in,clip,keepaspectratio]{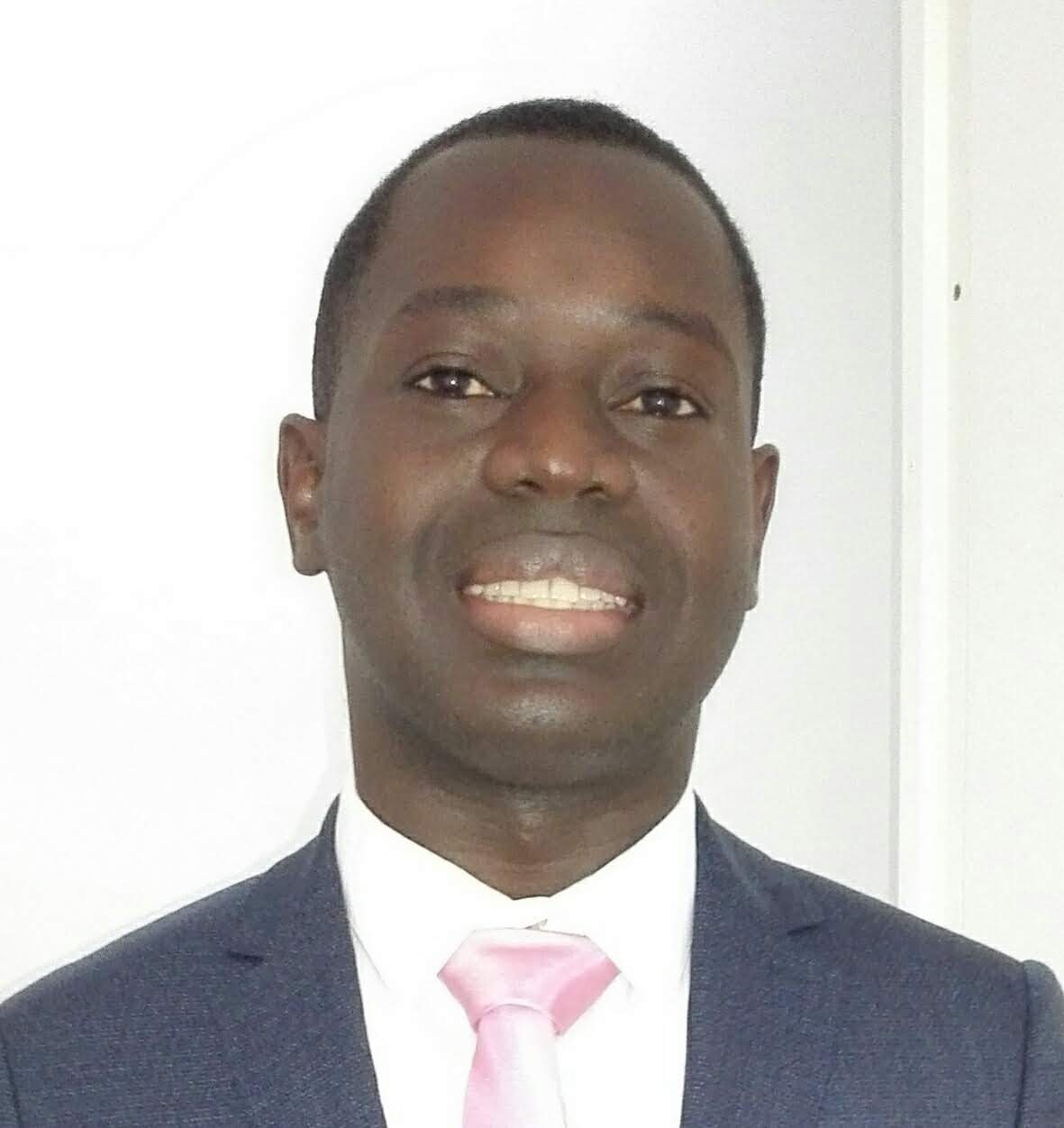}
  \end{wrapfigure}
  \textbf{D\'esir\'e Sidib\'e} received a Master degree from Ecole Centrale de Nantes and a Ph.D from Universit\'e de Montpellier in 2004 and 2007 respectively. From 2009 to 2019, he was an assistant, then associate professor at Universit\'e de Bourgogne and a member of the Le2i laboratory. Since 2019, he is a professor at Universit\'e Paris Saclay and a member of the IBISC laboratory. His main research areas include computer vision for autonomous vehicles, and medical image analysis. He has authored and co-authored more than 80 papers in international journals and conferences. He is a senior member of IEEE.\vspace{.1cm}\\
  
    \begin{wrapfigure}{l}{25mm} 
    \includegraphics[width=1in,height=1.25in,clip,keepaspectratio]{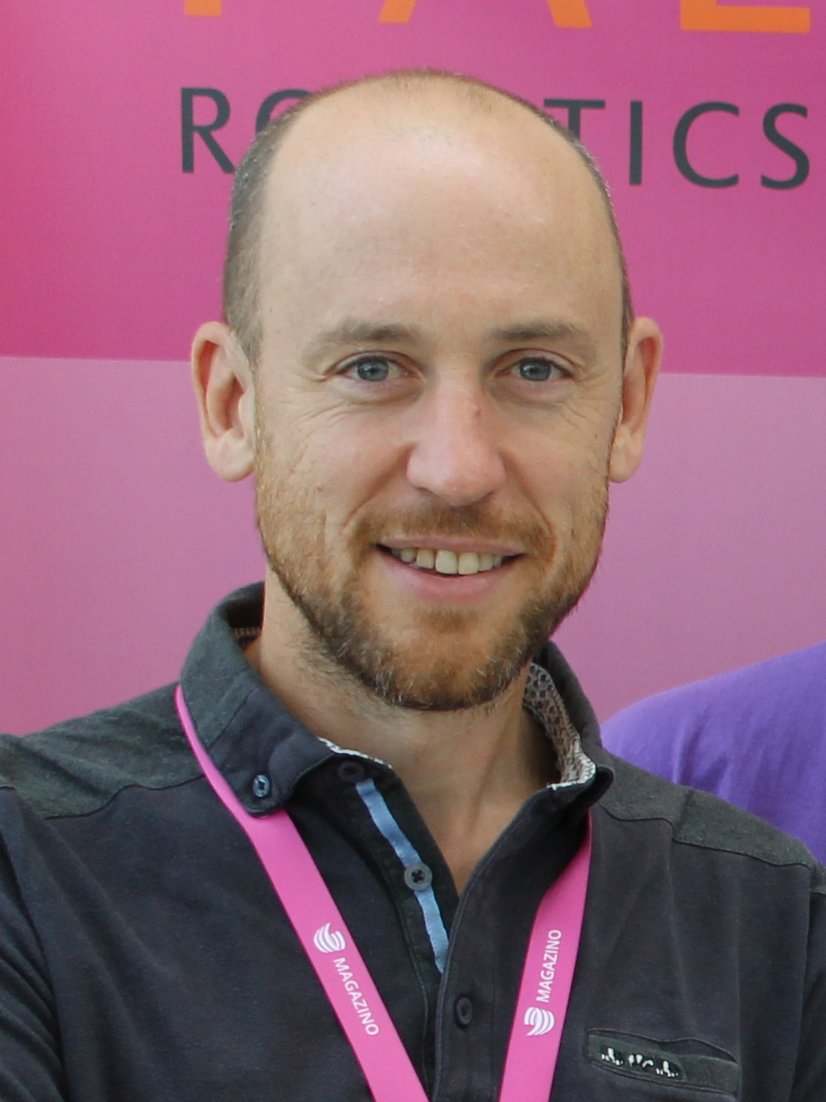}
  \end{wrapfigure}
  \textbf{Olivier Morel} received his MSc degree in computer vision and image processing from the University of Burgundy in 2002. In November 2005, he received his PhD degree in computer vision from the University of Burgundy. Since September 2007, he has worked as a lecturer in the VIBOT Team EMR CNRS 6000 from University of Bourgogne Franche-Comt\'e. His main research interests are polarization imaging and the applications of polarimetric vision to robotics and unmanned vehicles.\vspace{.1cm}\vfill
  
    \begin{wrapfigure}{l}{25mm} 
    \includegraphics[width=1in,height=1.25in,clip,keepaspectratio]{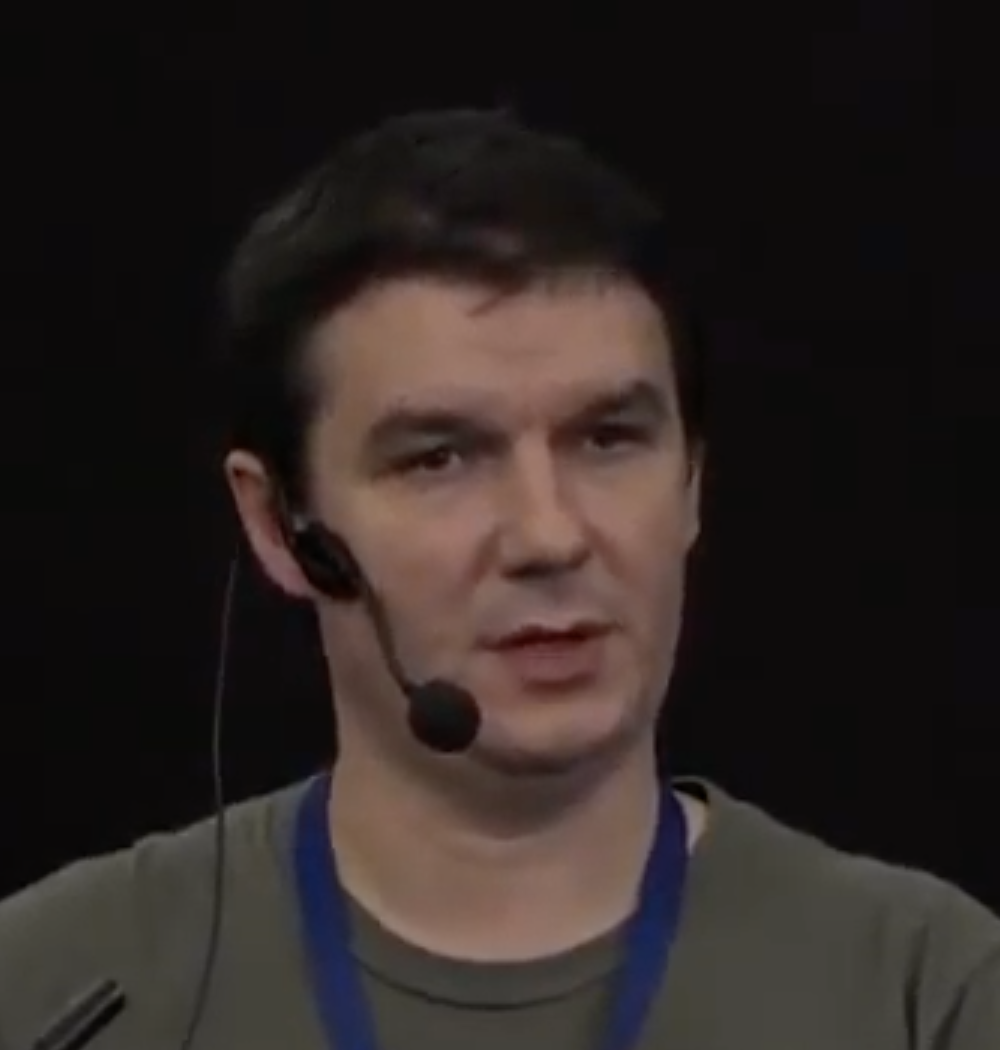}
  \end{wrapfigure}
  \textbf{Ralph Seulin} is a research engineer at CNRS in EMR VIBOT CNRS 6000 team. In 2002, he received his Ph.D degree in computer vision from the University of Burgundy. His main research areas include robotics, and computer vision for autonomous vehicles.
  He has authored and co-authored more than 80 international publications.\vspace{.1cm}\\
  
    \begin{wrapfigure}{l}{25mm} 
    \includegraphics[width=1in,height=1.25in,clip,keepaspectratio]{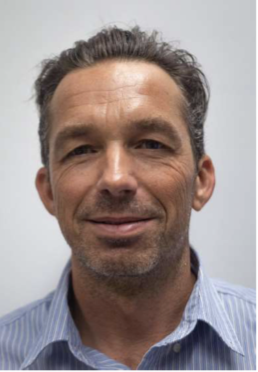}
  \end{wrapfigure}
  \textbf{Fabrice Meriaudeau} received both the master degree in physics at Dijon University, France as well as an Engineering Degree (FIRST) in material sciences in 1994. He also obtained a Ph.D. in image processing at the same University in 1997. He was a postdoc for a year at The Oak Ridge National Laboratory. He is currently Professeur des Universit\'es at the University of Burgundy.
He was the director of the Institute Health and Analytics (2017/2018) at the Universiti Teknologi PETRONAS Malaysia and was the Director of the Le2i (UMR CNRS) France, which has more than 200 staff members, from 2011 to 2016. His research interests were focused on image processing for non-conventional imaging systems (UV, IR, polarization) and more recently on medical/biomedical imaging. He coordinated an Erasmus Mundus Master in the field of Computer Vision and Robotics from 2006 to 2010 and was the Vice President for International Affairs for the University of Burgundy from 2010 to 2012. He has authored and co-authored more than 150 international publications and holds three patents.

\end{document}